# DroneAttention: Sparse Weighted Temporal Attention for Drone-Camera Based Activity Recognition

Santosh Kumar Yadav[a,b], Achleshwar Luthra[c], Esha Pahwa[c], Kamlesh Tiwari[c], Heena Rathore[d], Hari Mohan Pandey[e], Peter Corcoran[a]

[a]*College of Science and Engineering, National University of Ireland, Galway, H91TK33, Ireland*
[b]*CogniX, Quadrant-2, 10th Floor, Cyber Towers, Madhapur, Hyderabad, Telangana 500081, India.*
[c]*Department of CSIS, Birla Institute of Technology and Science Pilani, Pilani Campus, Rajasthan-333031, India*
[d]*Department of Computer Science at the University of Texas, San Antonio, United States.*
[e]*Department of Computing & Informatics, Bournemouth University, Poole, Dorset, BH12 5BB, United Kingdom*

## Abstract

Human activity recognition (HAR) using drone-mounted cameras has attracted considerable interest from the computer vision research community in recent years. A robust and efficient HAR system has a pivotal role in fields like video surveillance, crowd behavior analysis, sports analysis, and human-computer interaction. What makes it challenging are the complex poses, understanding different viewpoints, and the environmental scenarios where the action is taking place. To address such complexities, in this paper, we propose a novel Sparse Weighted Temporal Attention (SWTA) module to utilize sparsely sampled video frames for obtaining global weighted temporal attention. The proposed SWTA is comprised of two parts. First, temporal segment network that sparsely samples a given set of frames. Second, weighted temporal attention, which incorporates a fusion of attention maps derived from optical flow, with raw RGB images. This is followed by a basenet network, which comprises a convolutional neural network (CNN) module along with fully connected layers that provide us with activity recognition. The SWTA network can be used as a plug-in module to the existing deep CNN architectures, for optimizing them to learn temporal information by eliminating the need for a separate temporal stream. It has been evaluated on three publicly available benchmark datasets, namely Okutama, MOD20, and Drone-Action. The proposed model has received an accuracy of 72.76%, 92.56%, and 78.86% on the respective datasets thereby surpassing the previous state-of-the-art performances by a margin of 25.26%, 18.56%, and 2.94%, respectively.

*Email addresses:* santosh.yadav@pilani.bits-pilani.ac.in (Santosh Kumar Yadav), f20180401@pilani.bits-pilani.ac.in (Achleshwar Luthra), f20180675@pilani.bits-pilani.ac.in (Esha Pahwa), kamlesh.tiwari@pilani.bits-pilani.ac.in (Kamlesh Tiwari), heena.rathore@ieee.org (Heena Rathore), hpandey@bournemouth.ac.uk (Hari Mohan Pandey), peter.corcoran@nuigalway.ie (Peter Corcoran)

## 1. Introduction

Human Activity Recognition (HAR) is one of the emerging study fields in which human actions are determined based on the environment and the movement of body components. It has applications in numerous fields, including virtual reality, video surveillance, security, crowd behavior analysis, and human-computer interface. It comprises two main sub-tasks: classification and localization. While classification results in finding what a human is performing, localization refers to where the action is taking place in a scene of a video. Our study encompasses action classification using an efficient method to capture temporal data along with the spatial information in a single stream model while including operations to handle small objects. The model is trained and evaluated on three complex datasets, *i.e.*, Okutama-Action [1], Drone-Action [2], and MOD20 [3]. Figure 1 displays example images from these three datasets.

Human Action Recognition involves considerable obstacles that must be solved. Modern networks, such as 3D ConvNets, incur substantial processing costs. Training a 3D CNN backbone involves a substantial amount of time, slowing the search for an appropriate design and leading the model to overfit the training data. Aside from the length of model training, it is also important to remember that HAR lacks a defined lexicon of human actions. This can lead to the class's diversity. To solve this, it is necessary to build precise and distinguishing characteristics. Videos acquired from a distance, such as those captured by video surveillance cameras, are also a potential barrier to detecting the proper action because they cannot provide high-quality, clear images of the individual. The type of camera utilized also affects the performance recognition process. Real-time movies are visually dense and feature variable levels of brightness, making it difficult to discern activity in complex circumstances. In addition to background activities done by nearby objects, differences in scale, viewpoint, and partial occlusion can affect the model's output.

Owing to the recent surge in the literature on this topic, a large number of studies have been conducted on human activity recognition [4, 5, 6]. Multi-modal methods for HAR have gained popularity over the last few years. That being said, for a model to be successfully deployed for use, it has to be efficient and accurate. Due to these reasons, works such as PAN [7] and Temporal Shift Module (TSM) [8], which can be utilized with both 3D and 2D CNNs, were brought about. Various types of CNN architectures *e.g.* [9] and [10] implement two-stream networks of 2D-CNNs whereas works like [11] are an excellent example of 3D-CNN networks being both efficient and precise in their prediction. To learn long-range information from videos, Wang *et al.* proposed Temporal Segment

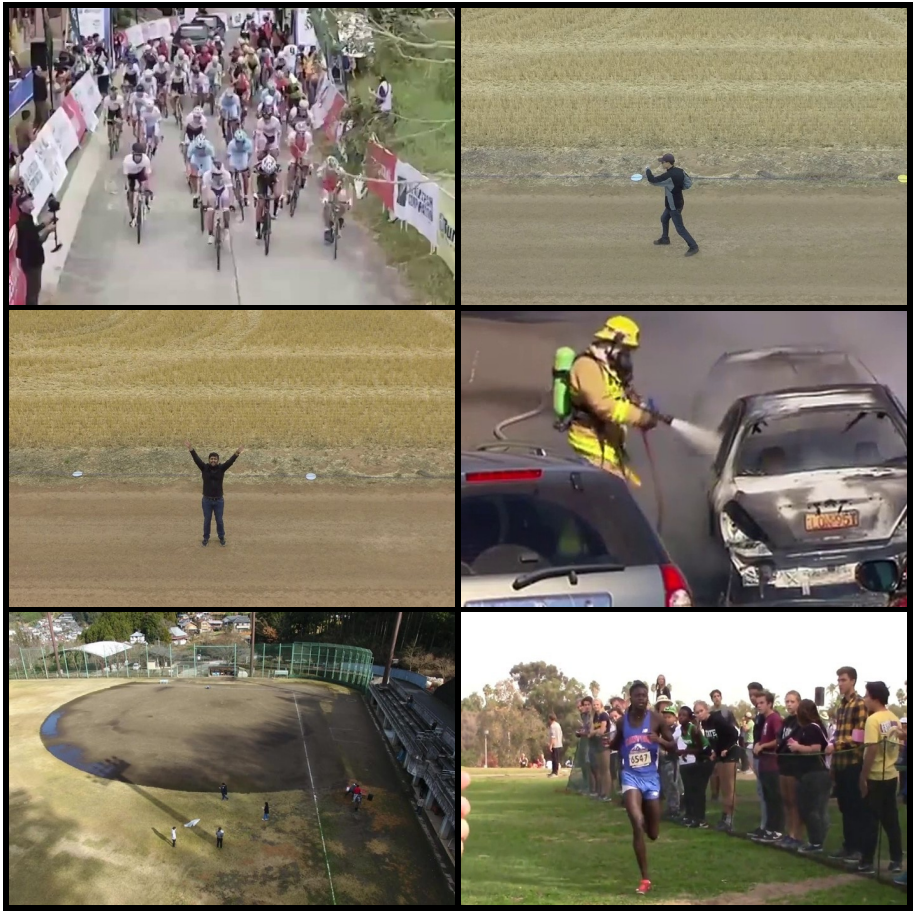

Figure 1: Examples of the three datasets used. The respective activities performed in the images are as follows (column-wise): 'cycling','punching', 'waving hands', 'fire-fighting', 'standing/walking', and 'running'.

Networks (TSN) [12] which uses segmented samples prior to feeding them to CNN architecture. Works such as Temporal Relation Network [13], Temporal Spatial Mapping [14], VLAD3 [15] and ActionVLAD [16], utilize this approach to deploy an efficient model.

The previous works mainly relied on a separate temporal stream to learn information available across a video. Our work tries to eliminate the necessity of utilizing a temporal stream by introducing a novel approach to fusing raw RGB frames with the optical flow in an efficient way. This research demonstrates the applicability of our plugin module for reducing computation costs by a huge margin. Thus, it makes drone-camera-based HAR more practical, fast, and coherent. Our method uses segment-based sampling [12] to include global temporal information with a minimum number of frames. We perform experiments on three diverse datasets and show that our method works better than the previous methods. Our SWTA (Sparse Weighted Temporal Attention) module can also boost the performance of existing approaches since it doesn't require knowledge about the internal details of the architecture such as activation functions, hidden layers, *etc*. and can be easily included in any method. The approach is easy to implement thus supporting faster experimentation. Other than that, the module can also act as a teacher network (the existing network being the student network), optimize the existing network to learn temporal information [17] and then it can be removed at the time of performance evaluation.

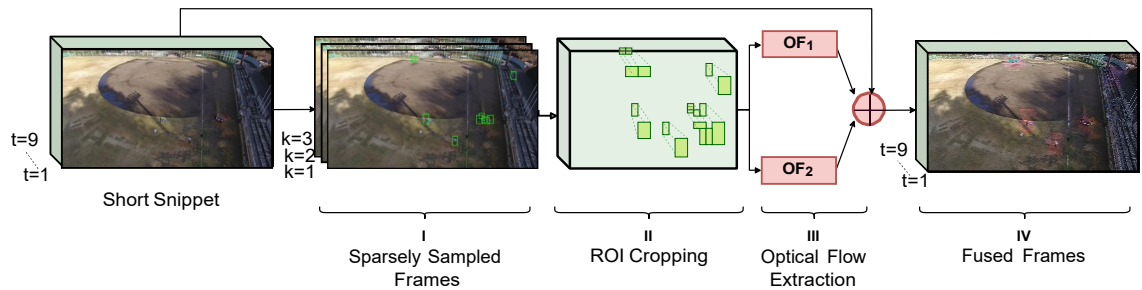

Figure 2: Detailed description of the SWTA (sparse weighted temporal attention) module. I) shows sparsely sampled frames obtained from a short snippet using a segment-based sampling technique. II) depicts how ROI(s) is(are) cropped across frames. III) and IV) illustrate optical flow extraction and their fusion with raw RGB frames respectively. Here values taken for *t* and *k* are for the demonstration purpose.

The major contributions of this paper are listed below:

- Firstly, sparse temporal sampling is introduced before the weighted temporal attention (WTA) module to obtain global attention with significantly lesser computation.

- Secondly, region of interest (ROI) cropping is incorporated in the WTA module to deal with the extremely small size (as shown in Figure 2) of human subjects. This helps in recognizing human activities from the high altitude of drone camera videos.

- Thirdly, we demonstrate how the proposed SWTA module acts as a plug-in module. Extensive computer simulations are performed on three publicly available benchmark datasets *i.e.*, Okutama dataset [1], MOD20 dataset [3], and Drone-Action dataset [2].

- Finally, the performance of the proposed model is determined, and a comparison of simulation results is performed with state-of-the-art methods. It is noted that the proposed model achieves state-of-the-art performance on the three datasets. Simulation results are organized in tabulated form in Table 3, 4 and 5 for quick review.

The rest of the paper is organized as follows: Section 2 provides a literature analysis of HAR and highlights its problems in adapting to drone-based HAR. Section 3 describes the proposed approach along with data preprocessing techniques. Section 4 offers implementation-specific information, such as dataset-specific information, experimental settings, evaluation measures, performance evaluation, visual results, and a comparison to other approaches. Section 5 is devoted to the concluding remarks and future directions.

## 2. Related Works

HAR is a vital topic of research that is being investigated in a variety of fields. As a result of recent developments in the field of deep learning and the data deluge, techniques for the task at hand continue to evolve and produce results that are state-of-the-art. This section highlights the existing scientific works conducted on the HAR to overcome the difficult task of identifying the actions of individuals in videos captured in different environments for a variety of purposes, including surveillance, sports analysis, crowd behavior analysis, and fall detection, among others. To provide a logical overview, discussions on linked works are grouped into four subsections. In conclude, we explain how the proposed approach addresses the issues posed by existing systems.

*2.1. Two-stream Networks*

Simonyan and Zisserman [18] proposed two-stream networks, namely, spatial and temporal streams, to achieve high accuracy in action recognition. This method was based upon the hypothesis [19] that the human visual cortex contains two pathways (a) the ventral stream (which performs object recognition); and (b) the dorsal stream (which recognizes motion). The spatial stream takes raw RGB frames as input whereas the temporal stream takes optical flow as input. The final prediction is the resultant average of both streams. But [18] used a relatively shallower network, as using a deeper neural network resulted in overfitting due to the unavailability of large video datasets. To overcome this issue, Wang *et al*. [9] had introduced a number of techniques including cross-modality initialization, synchronized batch normalization, corner cropping, multi-scale cropping data augmentation, large dropout ratio, *etc*. These techniques helped train deeper neural networks that surpassed [18] by a huge margin. Authors [20] had proposed pre-trained MotionNet, stacked before temporal-stream CNN, to generate optical flow, which was later fused with spatial stream CNN.

So far, the works discussed performs a late fusion of both the streams, *i.e.*, average the predictions of the spatial and temporal stream at the end of the network. Researchers have argued that this might not be the optimal way to fuse both streams. Kaparthy *et al*. [21] proposed different fusion methods such as early fusion, late fusion, and slow fusion for video classification. Authors [22] had shown that early fusion helps both streams learn richer features and leads to improved performance over late fusion. Feichtenhofer *et al*. [23] had introduced residual connections between two streams and further proposes a multiplicative gating function for residual networks to learn better spatio-temporal features. Researchers also explored recurrent neural networks (RNNs) to deal with videos. TS-LSTM [24] noticed that LSTMs require pre-segmented data to fully exploit the temporal information. Donahue *et al*. [25] introduced LRCN(Long term Recurrent Convolutional Networks) which combines convolutional layers with long-range temporal recursion but it gives a single prediction for the entire video and doesn't deal with multi-label HAR. Also, there was no clear improvement from RNN models over two-stream CNN models. Authors [10] had introduced VideoLSTM which uses spatial and motion-based attention mechanisms. It showed promising results and demonstrated the use of learned attention in action localization. Girdhar and Ramanan [26] extended attention maps to the existing architectures and provides factorization of attention processing into bottom-up saliency combined with top-down attention. Long *et al*. [27] argued that using long-term temporal information isn't always needed for video classification. Authors [27] had used multiple attention mechanisms units referred to as attention clusters to capture information from multiple modalities. In the attention action recognition network (AARN), the authors extracted spatial features from each frame in an action tube by using 2D convolutional layers. The extracted feature maps were then stacked together to form 3D feature maps. These features were fed to a spatio-temporal attention module (STAM), which acted as an autoencoder to generate attention maps that focus on the detected individuals.

Although these methods achieved high accuracy in learning short-range temporal information they were unable to capture long-range temporal information. To tackle this, Wang *et al*. [12] proposed temporal segment networks (TSN) that deal with capturing information available across long-range videos using segment-based sampling techniques and predict activities using five different segmental consensus functions like average pooling, max pooling, top-k pooling, linear weighting, and attention weighting. Although it was initially proposed to work with 2D CNNs, TSN being simple and generic but was used for 3D CNNs as well. Many existing works such as Temporal Relation Network [13], Temporal Spatial Mapping [14], VLAD3 [15] and ActionVLAD [16], that were proposed after [12] utilized the segment-based sampling in their approach.

*2.2. Multi-stream Networks*

Two-stream networks had performed well when given data was in the form of videos/image frames and optical flow, but other factors such as pose, depth, audio, *etc*. can also help in improving performance. In the RGB-D domain, Depth2Action [28] uses off-the-shelf depth estimators to extract depth information from videos and use it for action recognition. In [29] , the authors collected a multi-modal dataset combining data from 360° camera stream, LiDAR stream, and RGB-D stream captured by Depth cameras and achieved high accuracy but their dataset was limited to indoor activities. In [30], the authors argued on the importance of a representation derived from the human pose. They crop RGB image patches and flow patches for the right hand, left hand, upper body, full-body, and full image, based upon the joint estimations. Using these patches they used separate CNN architectures to extract their appearance and motion features which were further aggregated to provide video-level descriptors. Their model's performance was based on the accuracy of joints estimation which itself is a challenging task in high-resolution high-altitude videos that are collected by unmanned aerial vehicles (UAVs) because of various reasons such as unstable cameras due to wind, change in altitude, movement of UAVs, etc. Joint estimation in end-to-end training and prediction also adds extra computation which is not suitable for practical applications.

*2.3. 3-Dimensional CNNs*

3D-CNNs can serve as an alternative to optical flow methods and can be utilized as a processing unit to understand the temporal data present in the frames collected.

One of the first influential works on using 3D CNNs for action recognition was [31], but due to a shallow network, it could not capture all the spatial information provided. That gave rise to C3D [32] with a deeper 3D network that used a simple temporal pooling technique for action recognition, although it was unable to perform well on benchmark datasets as deep 3D CNNs are hard to optimize. That being said, it has since been frequently used as an optimal feature extractor for a number of video tasks. I3D introduced in [33] changed the outlook towards training 3D CNNs from scratch by adopting an architecture of stacked 3D convolution layers. It puts to use inflated ImageNet weights of 2D networks to their respective counterparts in the 3D network as proposed in [34]. I3D networks have also been used along with optical flow streams resulting in a two-stream model. As expected, it performed brilliantly on benchmark datasets [33].

To employ the benefits of pre-trained models on large datasets as already done for 2D-CNNs, Chen *et al*. [35] created ResNet3D by altering 2D into 3D filters. Approaches such as SENet [36] and STCNet [37] compute channel-wise data to adapt to both spatial and temporal features. Works on combining 2D and 3D CNNs to serve the interest of reducing the computational cost of deep 3D CNNs, have also been formulated. P3D network [38] factorizes 3D kernels into 2D and 1D kernels to better cope with the running complexity. Methods such as S3D [39] utilize the approach in [38] after replacing the bottom half of the 3D kernels with 2D kernels to generate a "top-heavy" network. The remaining 3D convolutions are factorized by P3D to further minimize the size of the model and reduce the time complexity. ECO [40] uses the same approach to get decent results for video understanding.

Long-range temporal modeling has also been operated on to get a better understanding of the change in scenarios over time. Stacking multiple temporal convolutions helps in performing this task. Although LTC [41] makes use of long-term temporal convolutions, resource issues such as limited GPU memory lead to feeding lower-resolution input images into the model which results in average results. Improvements were made in [42] as a non-local block was introduced. It captured both space and temporal domain information as it was placed after the residual blocks. Video classification was also performed by the CSN [11] efficient network which separates channel interactions and spatiotemporal interactions. SlowFast network [43] was proposed, which takes slow frame rate inputs and fast frame rate inputs separately into two streams. The former captures the semantic information while the latter operates at a high temporal resolution to learn the swiftly changing movements. However, this too is found to be more computationally expensive with respect to other architectures. A standard SlowFast network trained on benchmark datasets, on average takes 10 days to get completed.

*2.4. Efficient Architectures*

To eliminate the space cost of saving optical flow output separately on disk, [44] and [45] are among the earliest attempts to estimate the optical flow inside the training model. Unsupervised learning has also been applied to action recognition tasks to learn motion information in [20]. PAN [7] mimics the optical flow method to calculate the difference between consecutive feature maps, but this fails to capture long-term information. Approaches such as MARS [46] and D3D [17] use knowledge distillation mechanisms for combining two-stream networks into a single-stream network. This reduces the computational cost to a good extent but some amount of information is lost.

Both 3D and 2D CNNs approaches exist which used another efficient method called Temporal Shift Module (TSM) [8]. As the name suggests, it was used to shift part of the channels along the temporal dimension which promptly helped in information exchange among neighboring frames.

A number of advancements had been made for both indoor and outdoor datasets. The studies based on indoor datasets exploit RGB-D sensors or wearable sensing devices. These datasets include REALDISP gymnastic dataset [47], IM-dailydepthactivity dataset [48], MSRDailyAcitivity3D dataset [49], MuHAVi dataset [50], Charades dataset [51], and LBOROHAR dataset [29] among others. Outdoor datasets, include KTH dataset, VIRAT dataset [52], ViSOR dataset [53], Weizmann dataset, UCF-101 dataset [54] and Olympic Sports dataset among many others. A summary of the reviewed literature is presented in Table 1 for a quick review. In this study, we focus on three outdoor datasets, namely, the Okutama dataset, the Drone Action dataset, and the MOD20 dataset. These datasets pose sufficient complexity and scope which are required for obtaining trained models that can fit in real-world scenarios.

*2.5. Limitations*

The existing state-of-the-art on Drone-Action dataset [30] heavily relies on the correct predictions of the joint stream. The state-of-the-art on MOD20 dataset [3] uses a two-stream approach and depends on motion-CNN for its accuracy. The state-of-the-art on Okutama dataset [58] uses features computed by 3D convolution neural networks plus a new set of features computed by a binary volume comparison (BVC) layer. BVC layer comprises three parts: a 3D-Conv layer with 12 non-trainable (*i.e.*, fixed) filters, a non-linear function, and a set of learnable weights. Features from both the streams: 3D CNNs and BVC layer are concatenated and passed to a capsule network for fi-nal activity prediction. The proposed method achieves superior results on the Drone-Action dataset without separately utilizing pose-stream. On the Drone-Action dataset, we surpass prior state-of-the-art without a separate temporal stream because the SWTA module efficiently learns global temporal information using weighted temporal attention. Similarly, using the Okutama dataset, we attain the prior state-of-the-art although our model is significantly less computationally intensive since it does not require distinct streams of 3D CNNs to handle temporal information.

Table 1: Summary of the literature review.

| Ref. | Year | Datasets | Network | HMDB51 | UCF101 | Kinetics |
|---|---|---|---|---|---|---|
| [18] | 2014 | UCF-101, HMDB-51 | Spatial + Optical low assisted temporal stream | 59.4% | 88.0% | - |
| [21] | 2014 | UCF-101 | Slow-Fusion Network | - | 65.4% | - |
| [9] | 2015 | UCF-101 | Deep ConvNets with more data augmentation methods | - | 91.4% | - |
| [32] | 2015 | HMDB51, UCF-101, Kinetics | 3D ConvNets | 56.8% | 82.3% | 59.5% |
| [25] | 2016 | UCF-101 | CNN followed by LSTM (Image+Flow) | - | 82.34% | - |
| [10] | 2016 | HMDB51, UCF-101 | Convolutional LSTM + Action Tubes + Motion-based Attention | 52.6% | 82.1% | - |
| [22] | 2016 | UCF-101, HMDB-51 | Two stream network with fusion layers | 69.2% | 93.5% | - |
| [23] | 2017 | UCF-101, HMDB-51 | Temporal Residual Networks | 67.2% | 93.9% | - |
| [24] | 2017 | UCF-101, HMDB-51 | Temporal Segment LSTMs | 69.0% | 94.1% | - |
| [26] | 2017 | HMDB51 | Pose regularized attention pooling | 52.2% | - | - |
| [27] | 2017 | HMDB51, UCF-101, Kinetics | Multimodal Attention Clusters with Shifting Operation | 69.2% | 94.6% | 79.4% |
| [20] | 2018 | UCF-101, HMDB-51 | Spatial stream + Pre-trained MotionNet prior to temporal stream | 78.7% | 97.1% | - |
| [8] | 2019 | HMDB51, UCF-101, Kinetics400 | Temporal Shift Module for video understanding | 73.5% | 95.9% | 74.1% |
| [55] | 2019 | HMDB51, UCF-101, Kinetics400 | Channel-wise SpatioTemporal Module | 72.2% | 96.2% | 73.7% |
| [56] | 2020 | HMDB51, Kinetics400 | Motion squeeze module in two-stream network | 77.4% | - | 76.4% |
| [57] | 2020 | HMDB51, UCF-101, Kinetics400 | Motion excitation and Multi-temporal aggregation module | 73.3% | 96.9% | 76.1% |

## 3. Proposed Methodology

This section provides an overview of the proposed methodology. Herein is a comprehensive discussion of each component of the proposed model architecture. We explain how these components have been integrated for efficient human action identification using drone cameras. Our model utilizes a brief sequence of video frames that are preprocessed. Then, K frames are chosen by sparse temporal sampling. OpenCV is used to extract optical flow for K frames and then the weighted temporal attention (WTA) module is used to fuse optical flow with RGB frames. Then, using our backbone network, Inception-v3 with batch-normalization, features are retrieved from fused frames (RGB and Optical Flow). This is followed by the ROIAligning module, which is used to combine subject-specific features. These features are further flattened and transmitted to fully connected layers, which are then followed by max-pooling to produce action categorization at the individual level. The steps involved in the proposed model are further elaborated in the following subsections. Section 3.1 presents data preprocessing techniques. Section 3.2 describes a temporal segment network [12]. Details about our novel sparse temporal sampling-based weighted temporal attention module are given in Section 3.3. Section 3.4 discusses the backbone network, ROIAlign module, and certain modifications that are made on top of it.

*3.1. Data Preprocessing*

We have used three different datasets to verify the performance of our approach. Each dataset differs from the others in terms of actions, number of frames, frame rate, resolution of cameras used to record the videos, environment, camera motion, and even annotations. While the MOD20 dataset only comes with ground truth action labels, the Okutama dataset provides ground truth bounding boxes as well. The drone-Action dataset goes one step ahead and provides ground truth pose annotations along with bounding boxes and frames. We have predicted bounding boxes separately for MOD20 as the intermediate layers of our novel WTA module rely on bounding box coordinates. We utilize the joint annotations provided with the Drone-Action dataset in a separate pose stream for a fair comparison with previous state-of-the-art methods. Nonetheless, our model achieves competitive results even without a pose-stream.

We use data augmentation techniques, such as random cropping and horizontal flip to prevent our model from adversarial examples. We resize our images while maintaining the aspect ratio, for which the details are discussed in Section 4.2. All the images are rescaled before being fed to the model, and bounding box coordinates are normalized as well.

$$Final\ Image = \frac{Original\ Image}{255.0} \times [0.5, 2.0] \qquad (1)$$

*3.2. Temporal Segment Network*

As described in [this paper], dense sampling leads 2D ConvNets to overfit the training dataset since video frames are recorded densely and the content changes slowly, resulting in little temporal information. Instead of using every frame, we employ a computationally efficient technique [12] that expedites the training process. We combine sparse and global sampling approaches constructed with segment-based sampling to extract data from the complete snippet using a minimal amount of frames. The segment count is fixed, ensuring that the computational cost remains constant throughout all snippets.

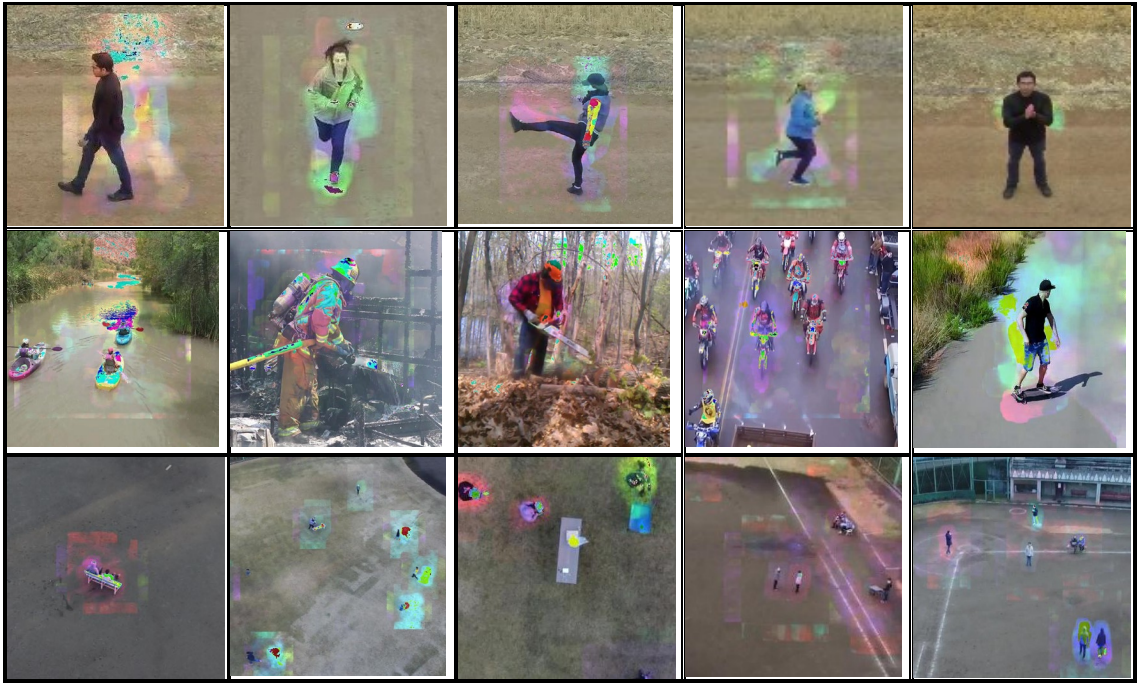

Figure 3: The given set of images depicts the action of weighted temporal attention of some selected frames. The first row contains examples from the Drone-Action dataset, of classes namely 'walking-sideways', 'running', 'kicking', 'running-sideways', and 'clapping' respectively. The second row shows examples from the MOD20 dataset of classes 'kayaking', 'fire-fighting', 'chainsawing-trees', 'motorbiking', and 'skateboarding' respectively. The third row shows examples from the Okutama dataset of various scenes where people can be seen sitting on a bench and interacting (first column); walking, standing (second column); carrying objects and interacting (third column); and standing (fourth and fifth column).

Given a short snippet $S$ whose shape is ($T, C, H, W$) where $T$ is the number of frames in a snippet, $C$ is a channel (e.g.: 3 for RGB), $H$ and $W$ are height and width respectively.

$$S = \{f_1, \ldots, f_T\}; \quad \text{where } f_i \in \mathrm{IR}^{(C \times H \times W)} \ \forall \ i \in [1, T] \tag{2}$$

where $f_i$ denotes $i_{th}$ frame in the snippet.

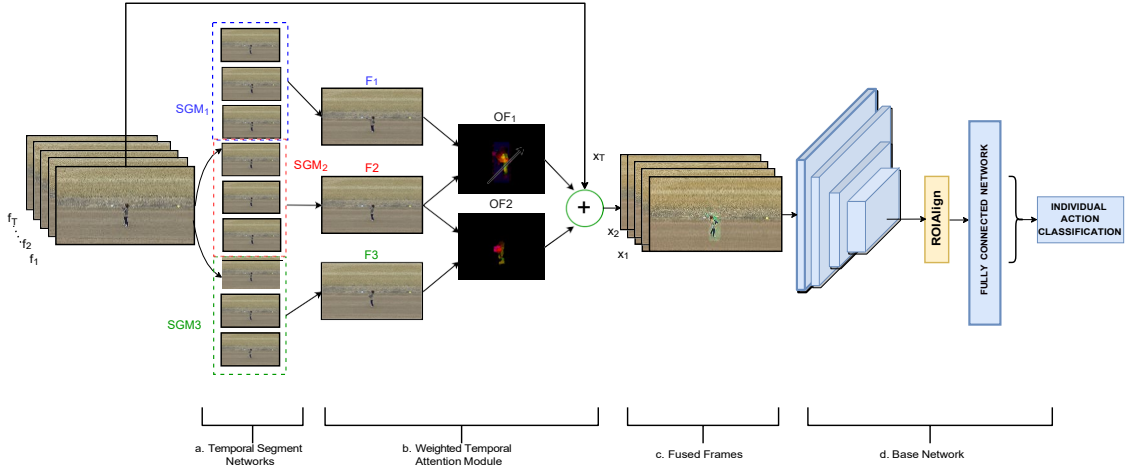

Figure 4: Block diagram of the DroneAttention. (a). A clip from the extracted frames is sparsely sampled and segmented into equal halves. (b). Random frames are chosen out of those segments and using the WTA module, we derive optical flow attention maps. (c). These maps are fused with the original frames. (d). The fused frames are fed into the base network which provides us with the activity classification results.

We divide it into $K$ segments $\{SGM_1, \dots SGM_K\}$ of equal durations and select one frame from each segment based on random sampling, as demonstrated in Figure 3.

$$SGM_i = \{f_{((i-1)\cdot k)+1}, \dots, f_{(i\cdot k)}\}; \qquad \forall\, (i \cdot k) \leq T,\ i \leq K \qquad (3)$$

We randomly select one frame F from each segment which implies:

$$F_i \in SGM_i \qquad \forall\, i \in [1, K] \qquad (4)$$

The use of a sparse sampling strategy reduces computational complexity dramatically and prevents overfitting which would have otherwise occurred due to a limited number of frames. Thus, it provides us with an efficient video-level framework that is capable of capturing long-range temporal structures.

### 3.3. Weighted Temporal Attention

We developed our novel Weighted Temporal Attention (WTA) module inspired by the extensive research on attention mechanisms in the past years [59, 60, 61]. WTA module takes sparsely sampled frames from Temporal Segment Network as input whose shape = ($K, C, H, W$). It learns attention maps that focus on specific parts Figure 2 of input relevant to the task in hand and our attention maps automatically lead to a sizable improvement in accuracy over baseline architectures.

Let $O(x, y)$ denote optical flow between $x$ and $y$, x and y being two frames:

$$OF_i = O(F_i, F_{i+1}) \tag{5}$$

Let $x_F$ denote the weighted temporal attention of snippet S:

$$x_F = \sum_{i=1}^{K-1} W_i \cdot OF_i \tag{6}$$

We perform eliment-wise multiplication of $x_F$ with all the T frames in our snippet S such that:

$$x_t := x_F \odot x_t; \quad \forall t \in [1, T] \tag{7}$$

In our work, we take a clip of T=15 frames. This snippet is divided into 3 segments of 5 frames each, and a random frame is sampled from each of these segments. Thus, we now get 3 frames - ($F_1$, $F_2$, $F_3$) from which optical flow is calculated: $OF_1 = O(F_1, F_2)$, $OF_2 = O(F_2, F_3)$. Each of these optical flow values is multiplied with $W_i$ as 0.033 and summation takes place ($x_F$). This value is then multiplied by each of the 15 frames.

The WTA module is a simple yet efficient module for integrating attention mechanisms into action recognition. Our unique method is easy to apply and may be viewed as an extension of "weighted average pooling." We include bounding box coordinates in the intermediate layers to help our novel WTA module to seek for relevant actions. We finally get an output of the shape - (1, C, H, W), which are further fused with RGB frames as shown in Figure 4. This approach helps us in reasoning for the long-term temporal relations even by looking at a single frame. We finally combine appearances from static RGB images and motion inferred by the WTA module to perform action recognition [62].

*3.4. Backbone Network*

The majority of past work, including cutting-edge algorithms, employs two-stream ConvNets in their design to handle appearance and motion independently. But the challenge is whether activities can be classified using a single stream of CNNs. In our technique, we merged optical flow features with static images and employed a single-stream of Inception-v3 [63] to predict action at the individual level. Pre-training the backbone on large-scale image recognition datasets such as ImageNet [64] has proven to be an effective solution when the target dataset lacks sufficient training samples [18]. We use Inception-v3 [63] with Batch Normalization pre-trained on ImageNet, as a backbone network, due to its balance between accuracy and efficiency.

Our model falls under the risk of overfitting due to a limited number of training samples. To prevent this, we have relied on various regularization techniques. Batch Normalization is able to deal with the problem of covariate shift by estimating the activation mean and variance within each batch to normalize these activation values. This also helps in faster convergence. Further, we add dropout layers between our last fully connected layers having a dropout ratio of 0.3 before the global pooling layer. We use the Adam optimizer with a weight decay parameter set to 1e-4 which adds L2 norm regularization. These techniques prevent the high risk of overfitting and help in the generalization of our network.

## 4. Experimental Results

This section provides the details of experimental results on three datasets. The complete architecture of the proposed network used is given in Figure 4. The datasets utilized in this experiment are discussed in Section 4.1. Section 4.2 outlines the experimental settings utilized during model training. Section 4.3 offers generic definitions of the employed evaluation metric and loss function. We have done a performance evaluation in Section 4.4. Section 4.5 compares the results of the proposed model with other algorithms, and Section 4.6 includes the complexity analysis of the model. Finally, Section 4.7 presents the visual results.

### 4.1. Dataset Details

Drone-camera-based human activity recognition is an emerging area in the field of computer vision research, and several researchers have made significant contributions to this field by releasing their experimental datasets. We have chosen the Drone-Action, MOD20, and Okutama datasets in order to evaluate the performance of our method and demonstrate that our model generalizes effectively over a wide variety of datasets. Table 2 summarizes these three datasets.

**Drone-Action:** Drone-Action [2] dataset was recorded in an outdoor setting using a free-flying drone to record 13 different actions. The dataset contains 240 high-definition video clips consisting of 66,919 frames. The number of actors in the dataset is 10 and they performed each action 5–10 times. All the videos are provided with 1920 $\times$ 1080 resolution. The average duration of each action was 11.15 sec. To add variation to the dataset, the volunteers included 7 males and 3 females. Apart from that, it adds variations in terms of the orientation, camera movement caused by drone/wind gusts, and the body shapes of the actors. The dataset comes with action labels, bounding boxes, and body joint estimations computed using the widely used pose estimator OpenPose. Videos were recorded using a GoPro Hero 4 Black camera with an anti-fish eye replacement lens (5.4 mm, 10 MP, IR CUT) and a 3-axis Solo gimbal. All videos in the dataset are in HD (1920 $\times$ 1080) format captured at 25 fps.

Table 2: Summary of the datasets used.

| Dataset Name | Classes | Number of Clips | Duration | FPS | Resolution |
|---|---|---|---|---|---|
| Okutama (2017) [1] | 12 | 43 | 60.00 s | 30.00 | 3840 x 2160 |
| Drone-Action (2019) [2] | 13 | 240 | 11.15 s | 25.00 | 1920 x 1080 |
| MOD20 (2020) [3] | 20 | 2324 | 7.40 s | 29.97 | 720 x 720 |

**MOD20:** MOD20 [3] is a 20 class-multi-viewpoint action recognition dataset wherein videos have been collected in an outdoor environment. Some videos were captured from drone and others from YouTube. It adds generality to the group of action recognition datasets and is more representative of real-world situations owing to the conditions the videos are captured in and its multiple viewpoint feature. Such a dataset also comes in handy for tasks such as video surveillance. A total of 2324 video clips were grouped into 20 classes, having both ground-level and aerial viewpoints. Six of them were captured from the drone while the rest were collected from YouTube. For each class, the average number of clips was 116. Having a mean clip length of 7.4 seconds, each video was re-sampled at 29.97 fps, after being cropped and resized into 720 $\times$ 720 pixels while the aspect ratio remained undisturbed. The selected actions for the given task were backpacking, chainsawing trees, cliff jumping, cutting wood, cycling, figure skating, fire fighting, jet skiing, kayaking, motorbiking, NFL catches, rock climbing, running, skateboarding, skiing, standup paddling, surfing, windsurfing (all outdoor), dancing and fighting (both indoor and outdoor).

**Okutama-Action:** Okutama-Action [1] dataset consists of 43 minute-long fully-annotated sequences with 12 action classes. Okutama-Action features many challenges including the dynamic transition of actions, significant changes in scale and aspect ratio, abrupt camera movement, as well as multi-labeled actors. This dataset provides a real-world challenge for multi-labeled actors where an actor performs more than one action at the same time. It consists of three main types of actions: human-to-human interactions such as hugging, and handshaking; human-to-object interactions such as reading, drinking, pushing pulling, carrying, and calling; and none-interaction such as running, walking, lying, sitting, and standing. This dataset has the dynamic transition of actions where, in each video, up to 9 actors sequentially perform a diverse set of actions which makes it very challenging and interesting.

*4.2. Experimental Settings*

On a machine equipped with an Intel Xeon processor, 12GB VRAM, and Nvidia Titan XP GPU, the model was trained for 80 epochs for each dataset. The model was compiled using the Pytorch backend. All the frames extracted from the video datasets were scaled to 420 × 720 and then normalized. Along with this, they were grouped into a batch size (B) of 2 while taking 15 frames (T) at a time. The resulting data had a shape of (*B, T, H, W, C*), where *H*, *W* and *C* denote height, width, and the number of channels of the frame, respectively.

Using the Inception-v3 backbone, the feature maps obtained were processed in the ROIAlign function, having a crop size (*K* ×*K*) which in our case is 5×5, to get our desired ROIs. The features thus obtained had a shape of (*B* × *T* , *N* , *D*, *K*, *K*), where *N* refers to the maximum number of boxes in a frame, 12 in the case of Okutama Dataset and 1 in case of Drone-Action dataset and *D* refers to the number of channels of the outputted features. The result was therefore flattened and fed into a fully connected (FC) layer having 512 units (*M* ) followed by a dropout layer, with the dropout ratio being 0.3, and a batch normalization layer. The output of the FC block having a shape of (*B* × *T*, *N*, *M*), was passed to the classifier which gave us the resulting probability.

A number of experiments were performed with different values of hyperparameters. The train-to-test split ratio was held constant at 80 : 20 for all datasets. Adam optimizer with an initial learning rate of $10^{-5}$, $β_1$ and $β_2$ with a value of 0.9 and 0.999, and a weight decay of $10^{-4}$ was found to be the most suitable optimizer for the given task of action recognition as well as to prevent overfitting. A learning rate scheduler was utilized in order to decrease the learning rate by one-tenth of its value after every 40 epochs. The one-hot encoded targets and predictions were fed into Binary Cross Entropy Logits Loss owing to its satisfactory usability to process softmax outputs of the last layer of the model.

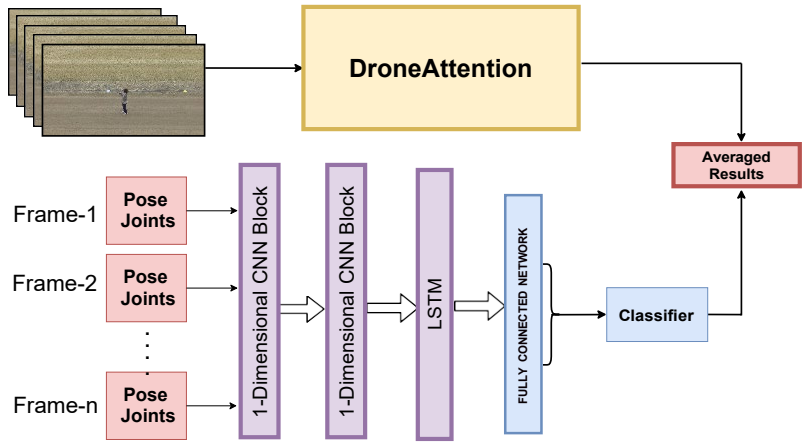

Figure 5: Block diagram of the pose-stream architecture integrated with our DroneAttention. As shown, the architecture includes two 1D CNN layers followed by an LSTM layer. The output is then processed in an FC layer, after which the averaged predicted results of activity recognition are obtained.

The Drone-Action dataset is specifically provided with pose annotations. Thus, it can be helpful to differentiate between similar classes such as jogging and running or hitting with a bottle versus hitting with a stick. A total of 18 joint coordinates (*x, y*) are provided for each frame in a JSON format. The normalized key points are grouped such that the final shape becomes (*B, T,* 18*,* 2), where *B* (*B* = 2) stands for batch size and *T* is the number of frames taken (*T* = 30). Each batch is thus fed into two 1D-CNN blocks, where each block corresponds to a 1D CNN layer followed by batch normalization and a dropout layer with a dropout ratio of 0.5. 1D CNN significantly helps in pattern recognition. The resultant features are passed to an LSTM containing 20 units, which measures the change in these features over the frames. Lastly, the softmax classifier is utilized to calculate the probability of flattened frame-wise features. The outcome of the function is fused with that of the Inception-v3 stream output by averaging the results and individual action classification is thus obtained. Using the Pytorch framework, Adam optimizer with $β_1$ and $β_2$ as 0.9 and 0.999 and a learning rate of 0.001 was used. Binary Cross Entropy With Logits Loss function was employed to calculate the loss of the one-hot encoded predictions. Figure 5 summarizes the entire architecture of the pose-stream module along with our proposed DroneAttention.

*4.3. Evaluation Metric*

Overall, the top-1 accuracy of the various model outputs for different datasets was chosen as an evaluation metric. Officially defined as the number of correct predictions over the total number of samples, for the respective categories, we use it as almost all classes in each dataset contain equal amounts of data. Hence, this metric is suitable for the given task for action recognition and classification.

$$Accuracy = \sum_{b=1}^{B} \sum_{t=1}^{T} \sum_{i=1}^{N} \frac{Correct\ Predictions}{Total\ Samples} \quad (8)$$

For each dataset, we get the shape of activity labels and bounding boxes as ($B, T, N$) where $B$ denotes the batch size, $T$, the number of frames per batch, and $N$ is the number of objects in each frame. We calculate the accuracy by computing the number of correct outputs for each frame for all $N$ objects. The resulting value is then averaged over all batches for each dataset to get our result.

*4.4. Performance Evaluation*

This section discusses the evaluation outcomes for the Okutama, MOD20, and Drone Action datasets. Table 3, 4, and 5 summarize the results of the respective datasets with the mentioned backbone that is utilized. Initially, we start our experiments using an Inception-v3 module [63] to capture the spatial features of the original images without the weighted temporal attention module. This helps us understand the critical and influential effect of the WTA module. Training the model using a basic backbone with RoiAlign, fully connected, batch-normalization, and dropout layers resulted in 61.34% accuracy for the Okutama dataset. This alone can be seen as outperforming the previous state-of-the-art values achieved using different backbones. To further improve the outcome, the proposed WTA module is added, and the backbone is fed with 'fused' frames instead of original ones, resulting in an overall accuracy of 72.76%, marking an increase of 11.42% from the baseline architecture and 25.26% from the previous state of the art. Using the optical flow backdrop, the network can specifically focus on the region where the action is taking place, and disregard the background which may contain noise.

Similarly for MOD20 dataset, consists of a diverse range of action classes, each significantly different from the other. It contains complex outdoor scenarios. That being said, our baseline model was able to achieve a higher accuracy of 90.03% as compared to the previous state-of-the-art value which was 74% [3]. After integrating the WTA module, a slight increase of 2.56% was obtained.

Table 3: Comparison with the state-of-the-art results on the Okutama dataset [1]. **Blue** represents the previous state-of-the-art. **Red** denotes the best results. Accuracies of previous works have been replicated from [58].

| S.No. | Method | Backbone | Accuracy |
|---|---|---|---|
| Past Work | AARN [65][58] | C-RPN + YOLOv3-tiny | 33.75% |
| | Lite ECO [40] [58] | BN-Inception + 3D-Resnet-18 | 36.25% |
| | I3D(RGB) [33] [58] | 3D CNN backbone | 38.12% |
| | 3DCapsNet-DR [66] [58] | 3D CNN + Capsule | 39.37% |
| | 3DCapsNet-EM [66] [58] | 3D CNN + Capsule | 41.87% |
| | DroneCaps [58] | 3D CNN + BVC + Capsule | **47.50%** |
| Ours | Baseline | Inception-v3 | 61.34% |
| | **DroneAttention** | **WTA + Inception-v3** | **72.76%** |

Table 4: State-of-the-art results on the MOD20 dataset [3]. **Blue** represents the previous state-of-the-art. **Red** denotes the best results.

| S.no. | Method | Backbone | Accuracy |
|---|---|---|---|
| Past | KRP-FS [67] | VGG-f + motion-CNN | **74.00%** |
| Ours | Baseline | Inception-v3 | 90.03% |
| | **DroneAttention** | **WTA + Inception-v3** | **92.56%** |

The Drone Action dataset contained various action classes which were similar to one another. For example, jogging from the front, back, and sideways was similar to running front, back, and sideways. It was critical to exactly locate the joint positions in order to determine which action was being performed. Hence, without the pose annotations, results were obtained from our simple baseline: 62.79% and integrated WTA module with baseline: 71.79%. Training the model along with pose joints in a separate stream led us to achieve greater results than the previous state-of-the-art, marking an increase of 2.84%.

In Table 3, 4, and 5, experiments have been conducted without the WTA module. In all cases, the model performs better with WTA module and the Inception-v3 architecture. In Table 5, an additional pose stream is also included which gives a better accuracy. The most important component of the network is the WTA module, in addition to which we observe a huge increase in the accuracy value for all three datasets.

*4.5. Discussion and Comparison*

Our approach outperforms the previously existing methods. It successfully achieves state-of-the-art results in all three datasets, namely 72.76% on the Okutama dataset, 92.56% on the MOD20 dataset, and 71.79% on the Drone Action dataset without pose-stream whereas 78.86% with pose-stream. For the Okutama dataset specifically, our Weighted Temporal Attention module with RoiAlign leads the network to focus on the key points where the action is currently taking place, and ignores the background noise, as opposed to the previously used 3D CNNs in [58]. It is also computationally less expensive, being a single stream network as compared to the approaches used in the previous works for MOD20 and Okutama dataset evaluation.

Table 5: Comparison with the state-of-the-art results on the Drone-Action dataset [2]. **Blue** represents the previous state-of-the-art. **Red** denotes the best results.

| S.no. | Method | Backbone | Accuracy |
|---|---|---|---|
| Past | MHT-MAF [68] | Multiple Human Tracking | 43.87% |
| | HLPF [69] | NTraj+ descriptors | 64.36% |
| | PCNN [30] | 'VGG-f' + Action Tubes | **75.92%** |
| Ours | Baseline | Inception-v3 | 62.79% |
| | **DroneAttention** | **WTA + Inception-v3** | **71.79%** |
| | **DroneAttention+Pose-Stream** | **WTA+Inception-v3+Pose-Stream** | **78.86%** |

Jhuang *et al*. [69] uses the HLPF approach which focuses on temporal and spatial information but ignores the additional data of the objects or props used in performing the action. Consequently, Cheron *et al*. [30] propose P-CNN which uses the two-stream network to process RGB patches and flow patches is able to surpass HLPF results. With our simple CNN-LSTM model that is decently able to distinguish between similar classes, to get results using pose data and computationally cheap temporal segment network to process ROI cropped regions, our model is able to produce better results in a shorter amount of time.

*4.6. Complexity analysis*

The Big-O computation time for the steps performed in WTA is:

- Division of T frames in the snippet into K segments - $O(1)$.

- Random sampling of one frame from each segment - $O(1)$.

- Calculation of optical flow - $O((K-1)d^2)$, taking dimensions of frames as $d \times d$.

- Pixel-wise addition of the optical flows generated - $O(d^2)$

- Multiplication of resultant weighted temporal attention with the rest of the frames in the snippet - $O(Td^2)$.

Hence, the total time complexity of the WTA module after removing the lower order terms becomes: $O(d^2((K - 1) + T + 1))$. Time complexity of one forward pass of the entire architecture consisting of l layers considering one snippet of T frames is fed to the model, is: $O(Td^2l)$ + Time complexity of WTA module = $O(Td^2l) + O(d^2((K - 1) + T + 1)) = O(d^2(Tl + (K - 1) + T + 1))$.

The Big-O space complexity for the extra arrays or matrices created to store the results of the steps performed in WTA is:

- Calculation of optical flow - $O((K - 1)d^2)$, taking dimensions of frames as $d \times d$.
- Pixel-wise addition of the optical flows generated - $O(d^2)$
- Multiplication of resultant weighted temporal attention with the rest of the frames in the snippet - $O(Td^2)$.
- As this module does not contain any learnable layers, the count of the number of parameters is zero.

Hence, the total space complexity of the WTA module becomes of the order: $O(d^2((K - 1) + T + 1))$.

The space complexity of the entire architecture consisting of the average number of parameters as p and number of layers as l when total frames (N) are fed to it, is: $O(p + l + Nd^2) + O(d^2((K - 1) + T + 1))$.

For most previous methods mentioned in Table 3, 4, and 5, due to the unavailability of official code, inference time could not be compared. The original manuscripts of the previous works focus on calculating the inference time for object detection and classification, which limits us to compare the time for classification alone.

*4.7. Visual Results*

Figure 6 and Figure 7 depict the correct and incorrect predictions made by our DroneAttention respectively. In Figure 6, the first row shows examples of the Drone-Action dataset [2], while the second and third-row show's examples of Okutama-Action [1] and MOD20 [3] datasets, respectively. The correct results are highlighted in green colored text whereas the incorrect ones are shown in red. A video showing the performance of the system during the experiments is available at https://youtu.be/9IogFm1BmZY.

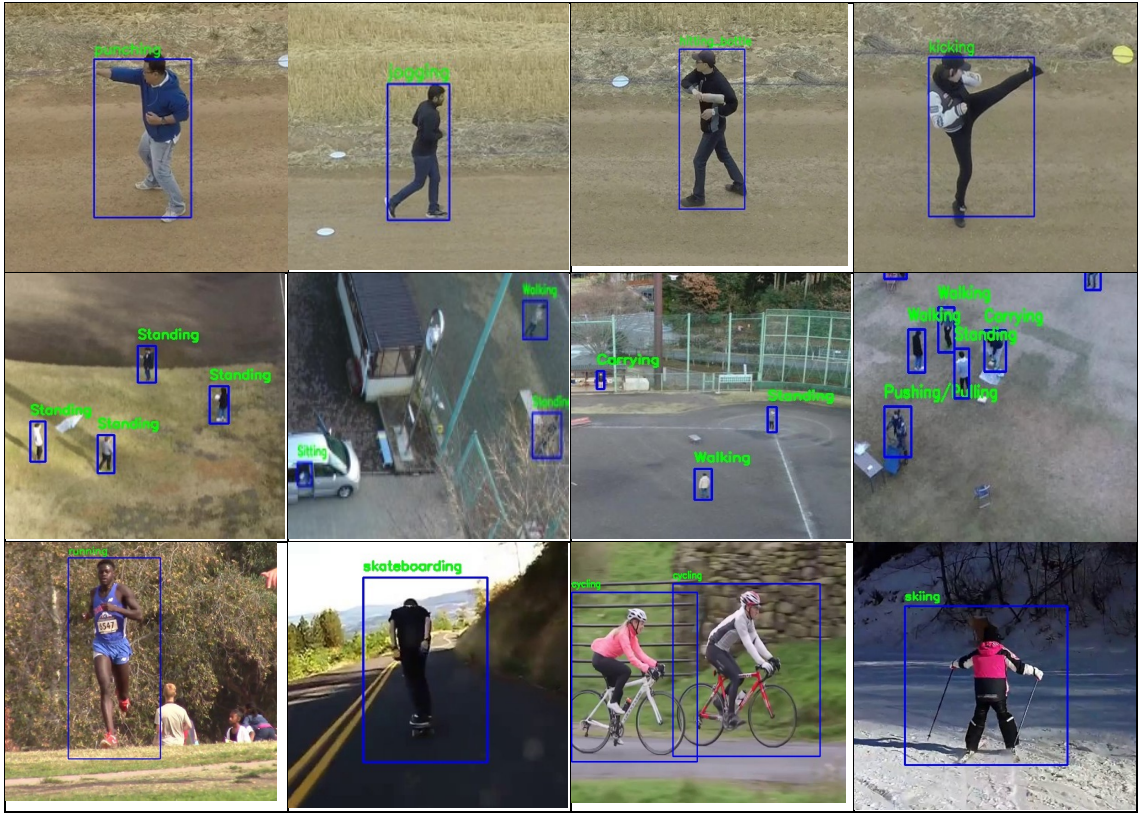

Figure 6: Examples of correct predictions made by our model on the three datasets, *i.e.* Drone-Action [2], Okutama-Action [1], and MOD20 [3]. A video showing the performance of the system during the experiments is available at https://youtu.be/9IogFm1BmZY.

## 5. Conclusion

In this paper, we present a DroneAttention network comprised of the Sparse Weighted Temporal Attention module that significantly improves our baseline's performance without significantly increasing its computing cost. We believe that our module can improve spatial streams for learning low-complexity temporal features by removing the requirement for a temporal stream, which is prevalent in activity recognition tasks involving deep learning. We have provided a novel method of integrating the concepts of temporal segment networks, weighted temporal attention, and con-volutional neural networks in order to determine the activity conducted in drone-camera-collected videos. Significant growth has been noticed in the Okutama-Action dataset, which is extremely valuable for drone-based activity recognition tasks at extremely high altitudes, such as crowd analysis and video monitoring. Our model achieves state-of-the-art performance on the challenging outdoor scenes depicted in the MOD20 and Drone-Action datasets, yet being less sophisticated than alternative approaches. Deploying such a model in a suitable technology could be more advantageous. This research has the potential to eliminate the computational obstacles preventing the adoption of deep learning-based HAR systems on drones.

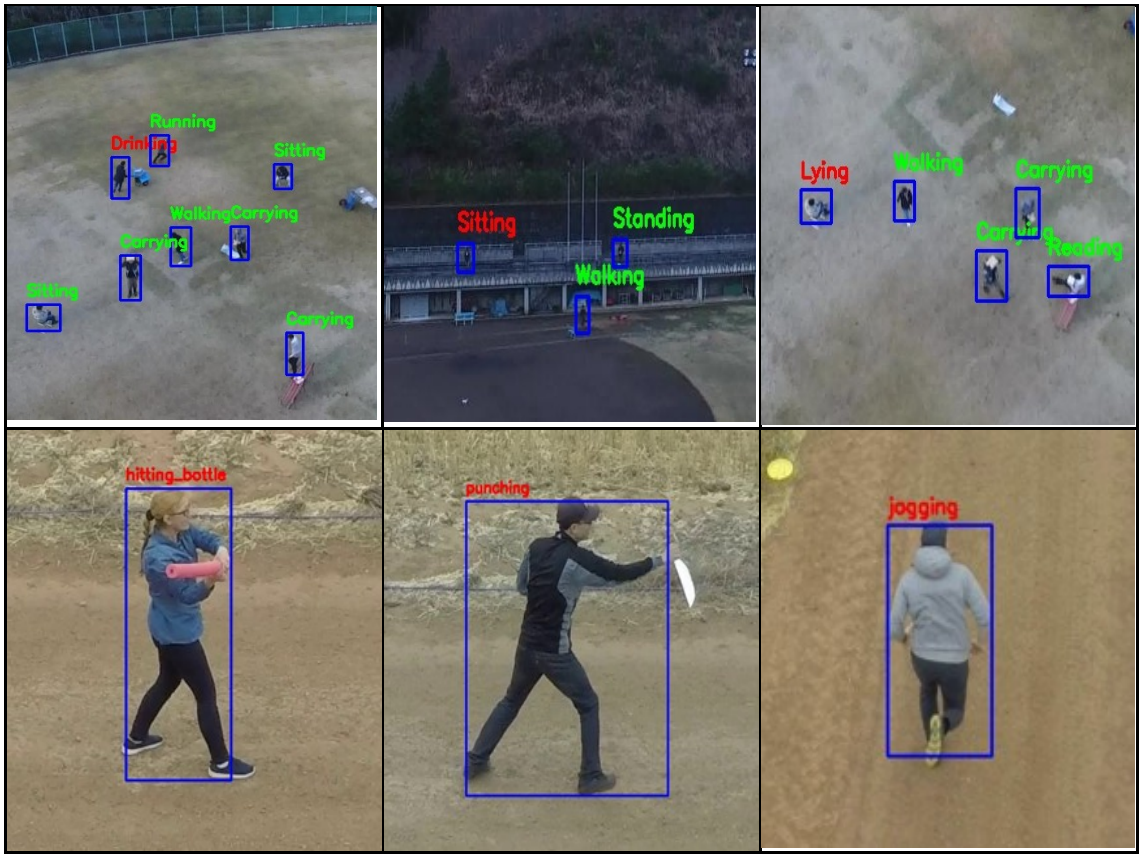

Figure 7: Examples of incorrect predictions made by our model on the Okutama-Action dataset [1] and Drone-Action [2]. In the first snapshot of the first row, a person annotated in red is seen 'carrying' an object, but the model predicts him to be 'drinking'; the second picture shows a person 'standing' but the model classifies him to be 'sitting'; the third snapshot, shows a person 'sitting' on a bench, but the model predicts him to be 'lying'. Ground truth values of the respective figures in the second row are: 'hitting stick', 'stabbing', and 'running'.

**Acknowledgment**

The authors would like to thank anonymous reviewers and our parent organizations for extending their support for the betterment of the manuscript. We appreciate the assistance provided by CSIR, India.

**Funding Information**

This research did not receive any specific grant from funding agencies in the public, commercial, or not-for-profit sectors.

**Conflict of Interest**

The authors declare that they have no conflict of interest.